\documentclass[DIV=calc, paper=letter, fontsize=11pt]{article}	 

\usepackage{lipsum} 
\usepackage{amsmath,amssymb,amsfonts,amsthm} 
\usepackage[english]{babel} 

\usepackage[colorlinks=true, linkcolor=blue, citecolor=blue, filecolor=blue, runcolor=blue, urlcolor = blue]{hyperref}

\usepackage{fullpage}
\usepackage{sectsty} 

\usepackage{fancyhdr} 
\usepackage{lastpage} 

\usepackage{nicefrac}
\usepackage{cite}

\usepackage{subcaption}

\usepackage{lettrine} 

\usepackage{caption}

\begin{document}

\thispagestyle{fancy} 

\noindent
{\small Position Paper at SciML2018 Workshop, US Department of Energy, January 2018. \\}\\
\textbf{\large Importance of the Mathematical Foundations of Machine Learning Methods \\for Scientific and Engineering Applications} \\ \textit{\large Paul J. Atzberger, Department of Mathematics, Department of Mechanical Engineering, \\ University of California Santa Barbara; atzberg@gmail.com; http://atzberger.org/}\\

There has been a lot of recent interest in adopting machine learning methods for scientific and engineering applications.  This has in large part been inspired by recent successes and advances in the domains of Natural Language Processing (NLP) and Image Classification (IC)~\cite{LeCun2015}.  However, scientific and engineering problems have their own unique characteristics and requirements raising new challenges for effective design and deployment of machine learning approaches.  There is a strong need for further mathematical developments on the foundations of machine learning methods to increase the level of rigor of employed methods and to ensure more reliable and interpretable results.  Also as reported in the recent literature on state-of-the-art results and indicated by the No Free Lunch Theorems of statistical learning theory incorporating some form of inductive bias and domain knowledge is essential to success~\cite{LeCun2015,Wolpert1996}.  Consequently, even for existing and widely used methods there is a strong need for further mathematical work to facilitate ways to incorporate prior scientific knowledge and related inductive biases into learning frameworks and algorithms.  We briefly discuss these topics and discuss some ideas for proceeding in this direction~\cite{Parrinello2007,KarniadakisLawrenceInfoFusion2017,KutzDiscoveryPDEs2017}.

In developing theory for machine learning methods it is important to  mention briefly the variety of modalities in which machine learning approaches are developed and deployed.  In \textit{supervised learning} one is interested in finding the functional relationship $f$ between labeled input data $x$ with output data as $y = f(x) + \xi$ often under non-ideal conditions such as with limited data, noise $\xi \neq 0$ or other uncertainties, or in large dimensional spaces.  Other modalities include \textit{unsupervised learning} with the aim is to discover inherent structure in the data and find a suitable parsimonious representation, \textit{semi-supervised learning} with partial labeling, or situations like \textit{reinforcement learning} where pay-offs change in response to actions.  We focus here on supervised learning but similar challenges arise for the other modalities.

It should be emphasized that the successes of many recent machine learning algorithms such as NLP and IC have hinged on clever use of prior knowledge concerning the nature of the data signal.  For instance in NLP the Word2Vec is used during a pre-training step to obtain embeddings of word identifiers into a space that encodes semantic similarities~\cite{LeCun2015}.  In IC use of Convolutional Neural Networks (CNNs) is widespread where prior knowledge about natural images are incorporated by the use of tied weights in the convolution filter banks to encode the importance of translation invariance~\cite{LeCun2015}.  This even includes intuition about the inherent hierarchical and compositional nature of such data signals in these problems which has motivated the current wave of \textit{deep learning} that favors in methods deep architectures over shallower ones in order to capture relevant information efficiently using distributed representations.

In the scientific and engineering domain there is a need for similar careful thought in order to gain domain specific insights and adaptations of machine learning algorithms so they can be effectively used and the full benefits of recent advances realized by the community.  To make more precise these ideas, we give one brief illustration for \textit{supervised learning}.  In contrast to traditional approximation theory, the goal is not only to approximate the solution $f$ from seen data but also hedge against uncertainty so as to obtain generalization of the model with good performance even for unseen data.  This can be captured by minimizing some notion of loss as captured by a function $\ell$ which defines the risk $R(f) = \mathbb{E}_{(x,y) \sim \mathcal{D}}\left[\ell(x,y,f)\right]$ such as in least-squares $\ell(x,y,f) = (f(x) - y)^2$ or as in maximum likelihood methods $\ell(x,y,f) = -\log(p(y|x,f))$.  However, $R(f)$ is seldom known in practice since one has limited information from data about the distribution $\mathcal{D}$ prompting the use of a surrogate error such as \textit{empirical risk} $\hat{R}(f) = \frac{1}{m}\sum_{i=1}^m\ell(x_i,y_i,f)$. As is well known to statisticians this requires caution since $\hat{R}$ may not converge uniformly to the true risk $R(f)$ as more data is acquired.  However, in the case of $f$ from a discrete hypothesis space $\mathcal{H}$ possibly infinite with any chosen notion of complexity $c(f)$ satisfying $\sum_{f\in\mathcal{H}} \exp(-c(f)) \leq 1$ one can derive a bound on the generalization error for $m$ samples as  
\begin{eqnarray}
R(f) \leq \hat{R}(f) + \sqrt{C_0\frac{c(f) + \log\left(\frac{1}{\delta}\right)}{m}}
\end{eqnarray}
which holds with probability $1 - \delta$ for randomly encountering data sets~\cite{Duchi2015}.  Similar bounds can be derived for continuous hypothesis spaces with other notions of complexity such as VC-dimension or Rademacher complexity.  This mathematically captures many current training protocols and learning algorithms which correspond to optimization of the RHS.  A common choice is \textit{empirical risk minimization} which for finite spaces corresponds to using $c(f) = \log(|\mathcal{H}|)$ where $c$ no longer plays a role in \textit{regularization}.

We can see how generalization can be achieved and performance improved by judicious choices of hypothesis spaces $\mathcal{H}$ and $c(f)$.  For scientific and engineering applications this could include incorporating prior information by designing $c(f)$ or restricting the space $\mathcal{H}$ to favor only functions that adhere to physical symmetries, satisfy constraints such as incompressibility, satisfy conservation laws, or more generally certain classes of linear or non-linear PDEs~\cite{KarniadakisLawrenceInfoFusion2017,KutzDiscoveryPDEs2017,Parrinello2007}.  This can have the abstract effect of better aligning good values of $c(f)$ and $\hat{R}$ and ensuring overall smaller risk $R(f)$.  While this traditionally has been a major emphasis in machine learning, this is not the only strategy.  As recent deep learning methods show, one can use complex hypothesis spaces but instead rely on the training protocols such as Stochastic Gradient Descent (SGD) that during selection favor lower complexity models in the sense of preserving only parts of the input signal $X$ relevant for prediction $Y$.  Similar opportunities exist for scientific and engineering applications where a lot of prior knowledge is often available about the relevant parts of input signals.  For example, as an alternative to limiting the hypothesis space, one could perform random rotations (group actions) on the input data during training to ensure the selected model makes predictions that are invariant under symmetries.  There are also many other possibilities combining these approaches using insights into the input data and ultimate aims.

We already see a number of insights can be obtained even for the type of generalization bounds we mention here which can be quite loose.  Further mathematical work on improving these bounds and analysis of training methods could be very beneficial in gaining further insights into how to effectively use existing methods or develop new approaches for incorporating prior knowledge.  We hope this brief position paper can serve as the start of the discussion on a few promising directions for mathematical investigations both in general theory and for characterization of current training algorithms in order to develop further frameworks and protocols better adapted for scientific and engineering applications.
\let\oldbibliography\thebibliography
\renewcommand{\thebibliography}[1]{%
  \oldbibliography{#1}%
  \setlength{\itemsep}{0pt}%
  \setlength{\parskip}{0em}
}
\bibliographystyle{plain}
\bibliography{paperDatabase}{}

\end{document}